\documentclass[agronomy,article,accept,moreauthors,pdftex]{mdpi} 
\firstpage{1} 
\pubvolume{xx}
\issuenum{1}
\articlenumber{5}
\pubyear{2019}
\copyrightyear{2019}
\usepackage{soul}
\history{Received: date; Accepted: date; Published: date}

\Title{Classification of Crop Tolerance to Heat and Drought—A Deep Convolutional Neural Networks Approach}

\Author{Saeed Khaki $^{1,}$*, Zahra Khalilzadeh $^{2}$ and Lizhi Wang $^{1}$}

\AuthorNames{Saeed Khaki, Zahra Khalilzadeh and Lizhi Wang}

\address{%
$^{1}$ \quad Department of Industrial Engineering, Iowa State University, Ames, IA 50010, USA; lzwang@iastate.edu\\
$^{2}$ \quad Institute of Transportation, Iowa State University, Ames, IA 50010, USA; zahrakh@iastate.edu}

\corres{Correspondence: skhaki@iastate.edu}

\abstract{Environmental stresses, such as drought and heat, can cause substantial yield loss in~agriculture. As~such, hybrid crops that are tolerant to drought and heat stress would produce more consistent yields compared to the hybrids that are not tolerant to these stresses. In the 2019 Syngenta Crop Challenge, Syngenta released several large datasets that recorded the yield performances of 2452~corn hybrids planted in 1560~locations between 2008 and 2017 and asked participants to classify the corn hybrids as either tolerant or susceptible to drought stress, heat stress and combined drought and heat stress. However, no data was provided that classified any set of hybrids as tolerant or susceptible to any type of stress. In this paper, we present an unsupervised approach to solving this problem, which~was recognized as one of the winners in the 2019 Syngenta Crop~Challenge. Our~results labeled 121 hybrids as drought tolerant, 193 as heat tolerant and 29 as tolerant to both~stresses.}


\begin{document}


\section{Introduction}

Environmental factors, especially temperature and precipitation, greatly influence the growth and development of crops~\cite{chen2012characterization} and subsequently lead to huge economic losses in agriculture~\cite{peng2004rice}. For~example, the~2012 drought and heat reduced grain yield by 21\% compared to the previous 5 years in the United~States~\cite{chen2012characterization, boyer2013us}. Climate models also suggest that negative impacts of drought and heat stress are likely to become even more frequent and severe in the future~\cite{cayan2010future,sanderson2011regional}.

As a major crop worldwide, the~yield performance of corn under heat and drought stresses has been extensively studied. Drought has long been known as a limiting factor to corn yields~\cite{hlavinka2009effect,heisey1999cimmyt}, the~impact of which depends on not only the severity but also the timing. Early season drought limits plant growth and development~\cite{chen2012characterization,shaw1983estimates}. Drought occurring at V8 to V17 stages significantly affects the ear size and kernel numbers~\cite{farre2006comparative,chen2012characterization}. Drought that occurs during silking stage considerably reduces the kernel weight, causing an average of 20\% to 50\% yield loss~\cite{schussler1991maize,chen2012characterization}. Heat stress at developmental stages of plant such as V8 can also cause significant yield loss~\cite{lobell2011nonlinear,chen2010role}. Lobell~et~al.~\cite{lobell2011nonlinear} found that each degree day spent above 30~$^{\circ}$C decreased the final yield by 1\% under optimal rain-fed~conditions. They~modeled the effect of weather on corn yields using a linear fixed-effects model based on three weather variables, namely the sum of growing degree days between 8~$^{\circ}$C and 30~$^{\circ}$C, the~sum of growing degree days
above 30~$^{\circ}$C and~precipitation. Lobell and Burke~\cite{lobell2010use} used regression analysis such as panel regression and time-series regression to examine the effects of growing average temperature and growing season total precipitation on the corn yield and found that an increase of 2~$^{\circ}$C in temperature would cause a greater yield loss than a 20\% decrease in precipitation. Badu-Apraku~et~al.~\cite{badu1983effect} analyzed the effect of temperature on corn yield using linear regression analysis and found that corn yield decreased by 42\% when mean daily temperature increased by 6~$^{\circ}$C. Severe heat stress can cause leaf firing, which~is a phenomenon that permanent tissue injury happens to developing leaves and injured tissues dry out later~\cite{chen2012characterization}. Early reproductive stages of corn can also be negatively impacted by moderate heat stress, which~decreases pollination rate, kernel set and~kernel weight, causing significant yield loss~\cite{cairns2013identification,cantarero1999night,cheikh1994disruption}. Combined drought and heat stress can cause greater yield loss than either stress can do alone. Lobell~et~al.~\cite{lobell2011nonlinear} analyzed more than 20,000 historical corn trials in Africa and found that heat stress caused additional 0.7\% yield loss when combined with drought~stress.

To improve the corn performance under environmental stresses, seed companies have started developing stress-tolerant corn hybrids to alleviate the negative effects of drought and heat stresses. Drought-tolerant (DT) hybrids have been developed through traditional plant breeding such as Pioneer Optimum $\text{AQUAmax}^{\text{TM}}$ and Syngenta $\text{Artesian}^{\text{TM}}$ \cite{10.3389/fpls.2016.01534}. Adee~et~al.~\cite{10.3389/fpls.2016.01534} studied the yield advantage of DT hybrids compared to non-DT hybrids using segmented regression analysis and found that DT hybrids yielded 5\% to 7\% more than non-DT hybrids in high stress environments, while maintaining a comparable yield potential in high yielding environments. Chen~et~al.~\cite{chen2012characterization} examined corn yield trials using a general linear model in a split block design and identified some heat tolerant corn inbred lines that demonstrated an enhanced tolerance to elevated temperatures. There has been less effort devoted to breading for heat stress and combined drought and heat stress~\cite{cairns2013identification}, even though there is evidence that the effect of drought stress at higher temperature is not equal to the sum of the effects of both stresses~\cite{barnabas2008effect,rizhsky2002combined}. Cairns~et~al.~\cite{cairns2013identification} used a linear mixed model to analyze the effects of environment and crop genotype on the corn yield and found that tolerance to combined drought and heat stress in corn was genetically different from tolerance to individual stresses and~their results identified several donors tolerant to combined drought and heat stress such as La Posta Sequia C7-F64-2-6-2-2 and~DTPYC9-F46-1-2-1-2.

In this paper, we designed a two-step approach using deep convolutional neural networks (CNNs) and regression analysis to classify corn hybrids as either tolerant or susceptible to drought stress, heat stress and~combined drought and heat stress. Compared with the aforementioned approaches for crop tolerance studies in the literature, which~were mostly linear regression analysis of crop yield response to the environmental components such as temperature and precipitation, the~proposed method first used deep convolutional neural networks to develop stress metrics (environmental index) which represent the amount of stress that corn hybrids would face in any particular environment across a growing season. The~concept of environmental index is also known as phenotypic plasticity, which~is the amount by which individual characteristics of a hybrid are changed across different environments~\cite{10.3389/fpls.2016.01534, chapman2008use, sadras2009phenotypic, bradshaw1965evolutionary}. After~extracting stress metrics, we regressed the yield of hybrids against each stress metric and classified the hybrids based on the slopes of the regression lines. The~slopes of the regression lines indicate the sensitivities of the hybrid yield with respect to the stresses. Small slopes indicate tolerant hybrids, which~have more stable yields across different environments and~large slopes indicate non-tolerant hybrids, which~have less yield adaptability~\cite{10.3389/fpls.2016.01534}. 

Deep learning methods are representation learning methods that can process data in raw format to automatically discover the representations needed for detection or classification~\cite{LeCun2015}. Deep neural networks are also known to be universal approximator functions, which~can represent almost any complex function~\cite{hornik1989multilayer, Goodfellow2016}. CNNs are methods that process data in the form of multiple arrays such as 1D (signals and sequences), 2D (images) and~3D (video). CNNs are composed of multiple convolutional layers and pooling layers, followed by few fully-connected (FC) layers. The~design parameters of CNNs usually include the number of filters, filter size, stride and~type of padding. A~filter is a set of learnable weights with which we convolve the input. The~stride is the amount by which the filter shifts. Padding is the process of symmetrically adding zeros to the input matrix to preserve the input~size.


Recent studies on crop stress classification using neural networks include the {following—Etminan}
~et~al.~\cite{etminan2019determining} used an artificial neural network with a single hidden layer to identify the best drought-related indices such as yield
stability index and stress susceptible index to predict yield performance of drought-tolerant durum genotypes. An~et~al.~\cite{an2019identification} proposed a deep convolutional neural network to classify corn drought stress based on images captured from the field every two hours throughout the whole day by digital cameras. Their proposed method classified the images captured from the field to three classes of optimum moisture, light drought and~moderate drought~stress. Deep~learning-based methods have also been used to predict crop performance (such~as~yield) across different environments~\cite{kim2019comparison,wang2018deep,saeedCrop}, however, none of these studies examined whether hybrids were tolerant to environmental stresses such as heat and drought. Compared to our proposed method, which~is an unsupervised approach, all aforementioned methods are supervised approach. The~proposed unsupervised method can be used for clustering hybrids based on their stress tolerance when lack of data on the tolerance or susceptibility of hybrids prevents direct application of supervised models for the classification task. The~two-step nature of the proposed model is also advantageous for two main reasons—(1) it develops stress metrics which show how stressful each environment is compared to the other environments, (2) it makes the model less of a black box and more~expansible. {Mechanistic~models are an alternative approach to analyzing crop response to environmental stresses, which~rely on abundant experimental data to reveal the biological processes that have given rise to the crop response}{ \cite{hilborn1997ecological, bolker2008ecological}.{ In practice, when only sparse datasets are available for large numbers of hybrids and environments, machine learning models are more appropriate for estimating the complex relationship between explanatory and response variables to best fit the data.} Table~\ref{tab:lit} compares the crop stress tolerance studies in the literature with respect to their method, input data and~model~output.

\begin{table}[H]
\caption{Comparison of crop stress tolerance studies in the~literature.}\label{tab:lit}

\centering
\scalebox{0.88}{
\begin{tabular}{ccccc}

\toprule
\textbf{Literature} & \textbf{Method} & \textbf{Type}& \textbf{Input Data} &\begin{tabular}[x]{@{}c@{}} \textbf{Model Output}\end{tabular} \\
\midrule
Lobell~et~al.~\cite{lobell2011nonlinear}& linear fixed-effects model& supervised&weather & \begin{tabular}[x]{@{}c@{}} analysis of yield response to\\ heat and drought stresses\end{tabular}\\
\midrule

Lobell and Burke~\cite{lobell2010use}& \begin{tabular}[x]{@{}c@{}} panel regression and\\ time-series regression\end{tabular}& supervised &weather& \begin{tabular}[x]{@{}c@{}} analysis of yield response to\\ heat and drought stresses\end{tabular}\\
\midrule
Badu-Apraku~et~al.~\cite{badu1983effect} & linear regression model & supervised &weather& \begin{tabular}[x]{@{}c@{}} analysis of yield response to\\ heat stress\end{tabular}\\
\midrule
Adee~et~al.~\cite{10.3389/fpls.2016.01534}& segmented regression model& supervised &weather& \begin{tabular}[x]{@{}c@{}} analysis of yield response to\\ drought stress\end{tabular}\\
\midrule
Chen~et~al.~\cite{chen2012characterization}& general linear model&supervised & weather& \begin{tabular}[x]{@{}c@{}} analysis of yield response to\\ heat and drought stresses\end{tabular}\\
\midrule
Cairns~et~al.~\cite{cairns2013identification}& linear mixed model&supervised &weather & \begin{tabular}[x]{@{}c@{}} analysis of yield response to\\ heat and drought stresses\end{tabular}\\
\midrule
Etminan~et~al.~\cite{etminan2019determining} & artificial neural network &supervised &weather & \begin{tabular}[x]{@{}c@{}} identification of \\drought-related factors\end{tabular}\\
\midrule
An~et~al.~\cite{an2019identification} & CNN &supervised&digital image &\begin{tabular}[x]{@{}c@{}} classification of crop\\ tolerance to drought\end{tabular}\\
\midrule
Proposed method & CNN and linear regression&unsupervised& weather and soil& \begin{tabular}[x]{@{}c@{}} (1) classification of crop\\ tolerance to drought, heat\\ and their combination \\(2) stress metrics extraction\end{tabular}\\
\toprule
\end{tabular}
}

\end{table}

The remainder of this paper is organized as follows. Section~\ref{S2} defines the problem statement. Section~\ref{S3} presents the methods and results. Finally, we conclude the paper in Section~\ref{S4}.

\section{Problem~Statement}\label{S2}


In the 2019 Syngenta Crop Challenge~\cite{s2}, participants were asked to use real-world data to develop stress metrics and use these stress metrics to classify corn hybrids as either tolerant or susceptible to drought stress, heat stress and~combined drought and heat stress. {The underlying research problem is to identify whether a crop hybrid is tolerant to environmental stresses such as drought, heat and~combined drought and heat}. 

The data used in this research included four sets—performance, management, soil and~weather; no genotype data was provided to complement these four datasets. The~performance dataset contained the observed yields of 2452 experimental hybrids planted in 1560 environments. 



The management dataset contained planting/harvest dates and irrigation type. The~soil data included field elevation, percentage of clay, silt and sand, available water capacity, soil pH, organic matter, cation-exchange capacity and~soil conductivity. The~weather data included daily record of seven weather variables, namely day length, precipitation, solar radiation, snow water equivalent, maximum temperature, minimum temperature and~vapor~pressure.



 The irrigation type variable had 2.14\% missing values. Multiple imputation techniques were tried including median and most frequent and we found that the most frequent approach led to the most accurate result. The~irrigation type variable was converted into one-hot encoding format. Planting/harvest dates play an important role in the amount of stress that corn would face due to affecting other variables such as temperature and precipitation~\cite{nafziger1994corn}. Different environments had different planting/harvest dates, which~resulted in different growing season lengths. {Figures} \ref{fig:map_p} and \ref{fig:map_h} {show the maps of planting dates and harvest dates of environments across the United States}. Figure~\ref{fig:Growing_l} shows the histogram of growing season length across all~environments.

\begin{figure}[H]
\centering
\includegraphics[scale=0.25]{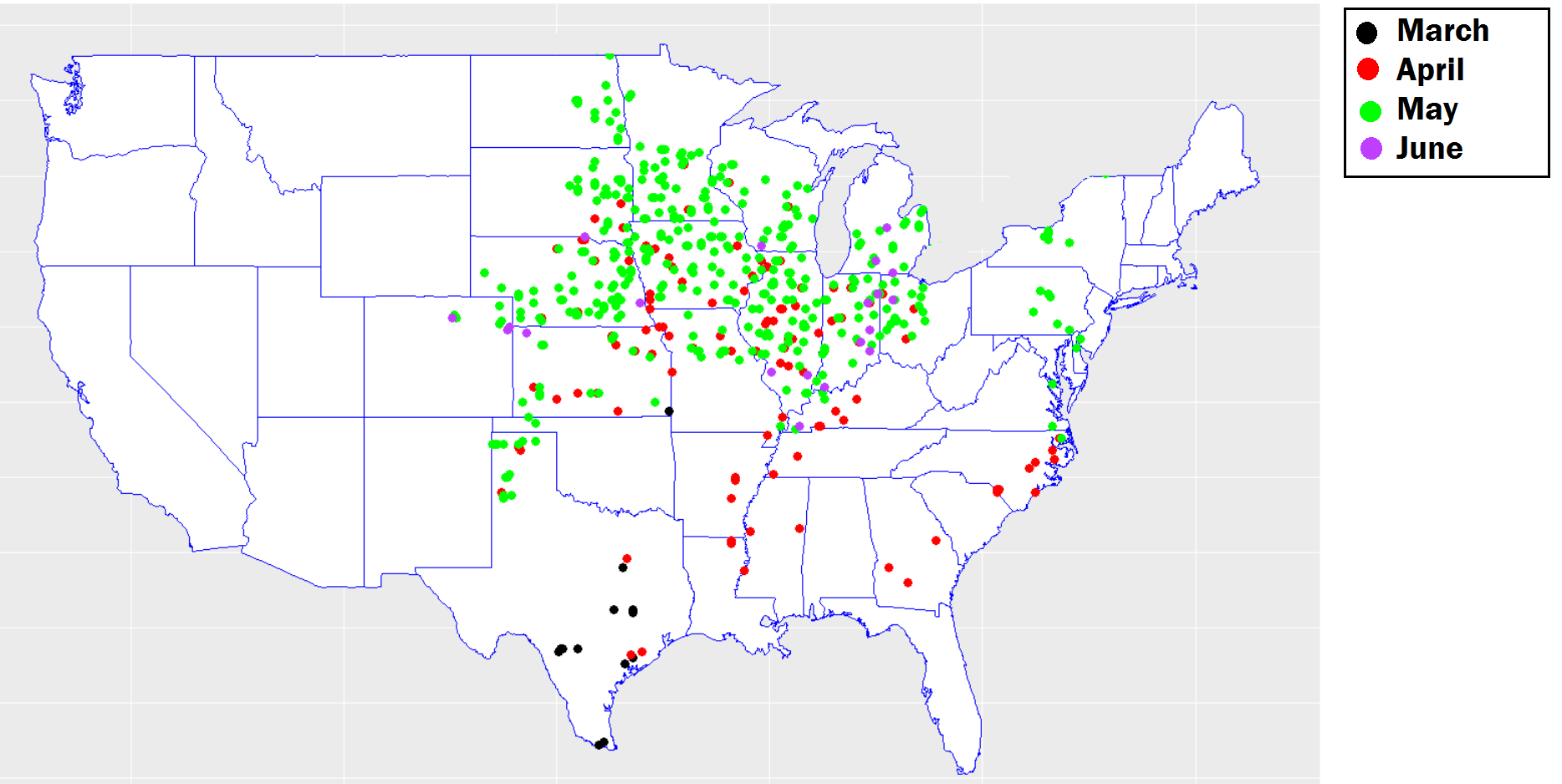}
\caption{{Map of planting dates of environments across the United States. Data collected from the 2019 Syngenta Crop Challenge} \cite{s2}.}\label{fig:map_p}

\end{figure}

\vspace{-6PT}

\begin{figure}[H]
\centering
\includegraphics[scale=0.25]{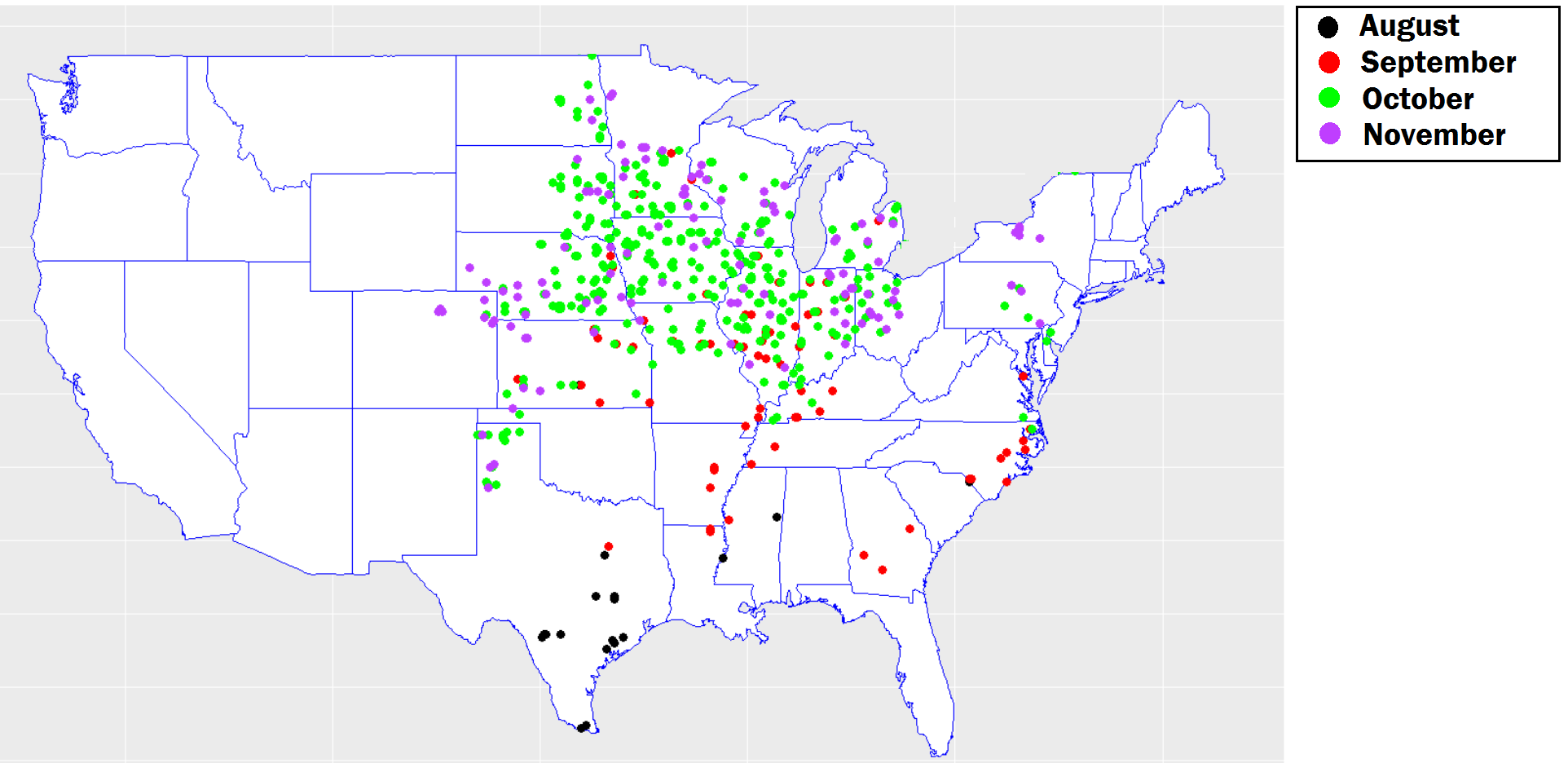}
\caption{{Map of harvest dates of environments across the United States. Data collected from the 2019 Syngenta Crop Challenge} \cite{s2}.}\label{fig:map_h}

\end{figure}
\unskip

\begin{figure}[H]
\centering
\includegraphics[scale=0.21]{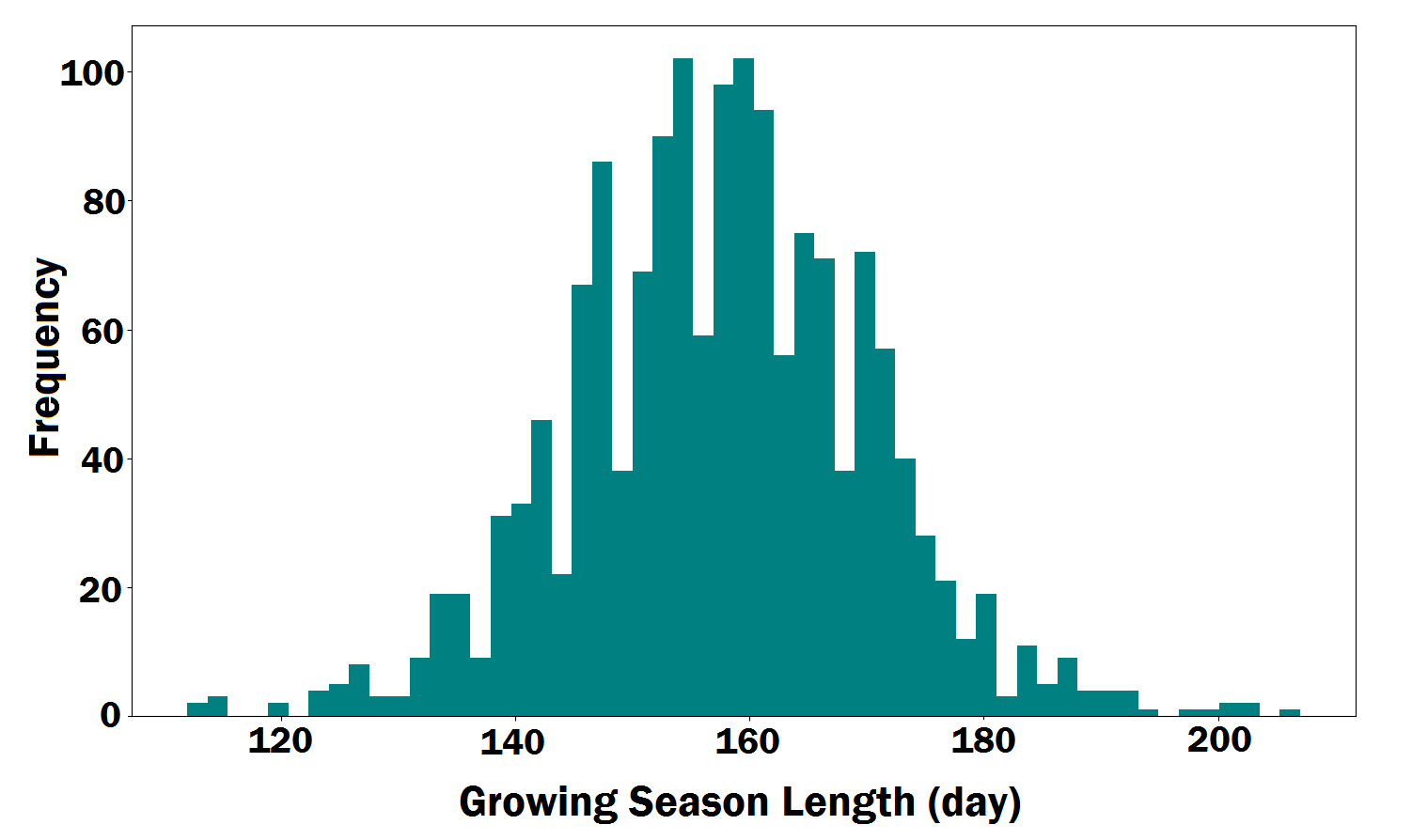}
\caption{Histogram of growing season length across all~environments.}\label{fig:Growing_l}

\end{figure}
To capture the effect of planting/harvest dates, we only used weather data between the corresponding planting/harvest dates for each environment. We divided the growing season for each environment into 20 intervals with equal length and~took the average of the daily weather records in each interval for each weather variable. For~example, the~growing season length of the environment 7 was 180 days, thus there were 9 days (180/20) in each interval for this environment. As~such, we had a new representation of the weather data which considered the the effect of planting/harvest dates while having equal number of features (20 features for each weather variable) for each environment. We tried other number of intervals and found that 20 intervals led to the most accurate results. Figure~\ref{fig:Growing2} shows the plots of weather variables for environment 7 over growing season using 20-interval representation. We used all variables, including weather in 20-interval representation format, soil and~irrigation type to train our proposed~model.
\vspace{-6pt}

\begin{figure}[H]
\centering
\includegraphics[scale=0.230]{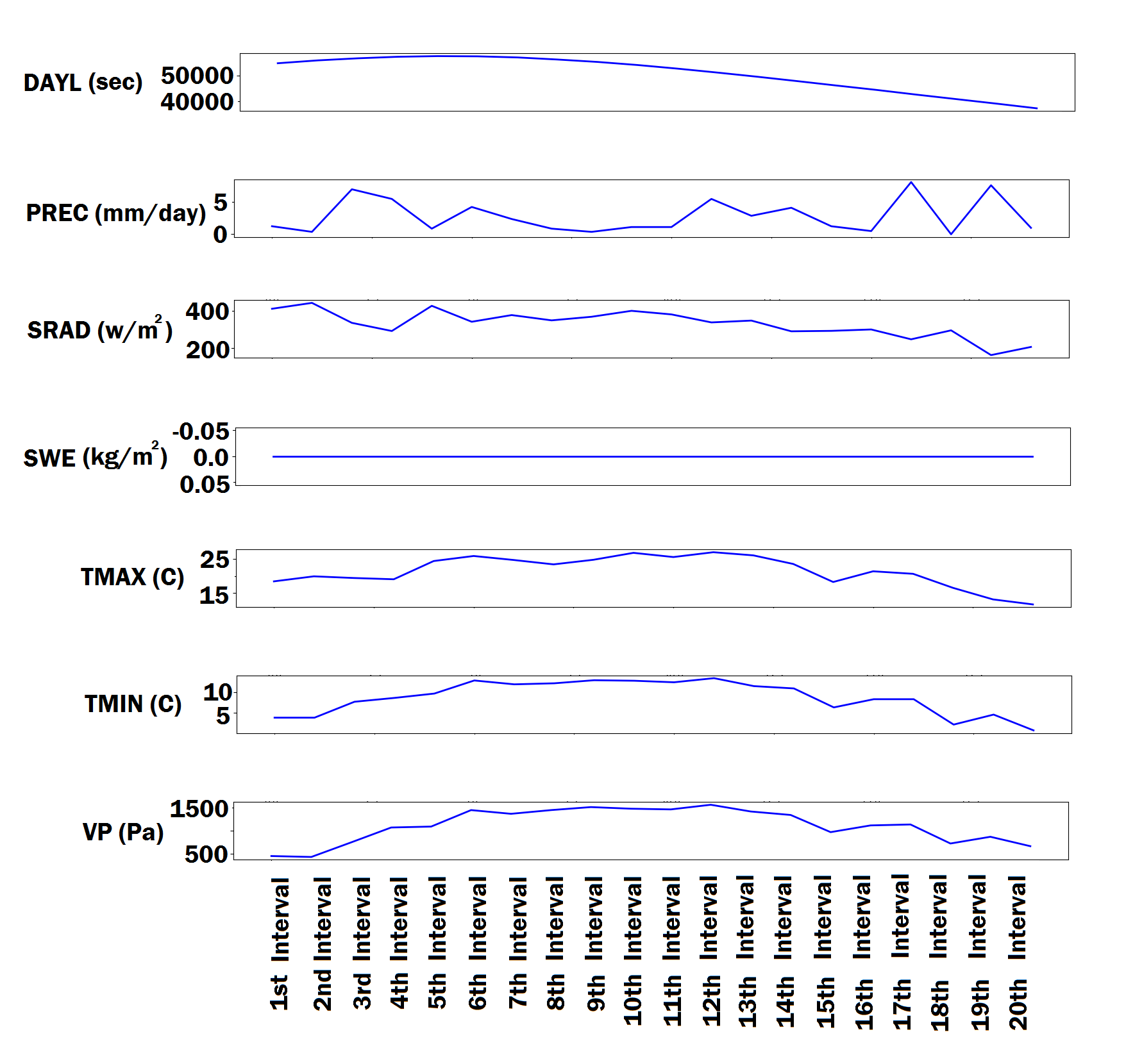}
\caption{Plot of environment 7's weather variables over growing season using 20-interval representation of data. DAYL, PREC, SRAD, SWE, TMAX, TMIN and~VP stand for day length, precipitation, solar radiation, snow water equivalent, maximum temperature, minimum temperature and~vapor pressure, respectively. { sec, mm/day, w/$\textit{m}^2$, kg/$\textit{m}^2$, C and~Pa stand for second, millimeters per day, watt per square meter, kilogram per square meter, Celsius and~Pascal, respectively.}}\label{fig:Growing2}

\end{figure}

\section{Methods and~Results}\label{S3}

We designed an unsupervised two-step approach, as~illustrated in Figure~\ref{fig:model}, to~the crop tolerance classification problem {since no data was provided that classified any set of hybrids as tolerant or susceptible to any type of stress.} In the first step, we train a deep CNN model to predict yield stress, which~is defined using historical yield data and~then we extract stress metrics using the trained CNN~parameters. In~the second step, we use regression analysis to classify corn hybrids as either tolerant or susceptible to the stress metrics extracted in the first step. The~following subsections describe the proposed method in more~detail.

\begin{figure}[H]
\centering
\includegraphics[scale=0.45]{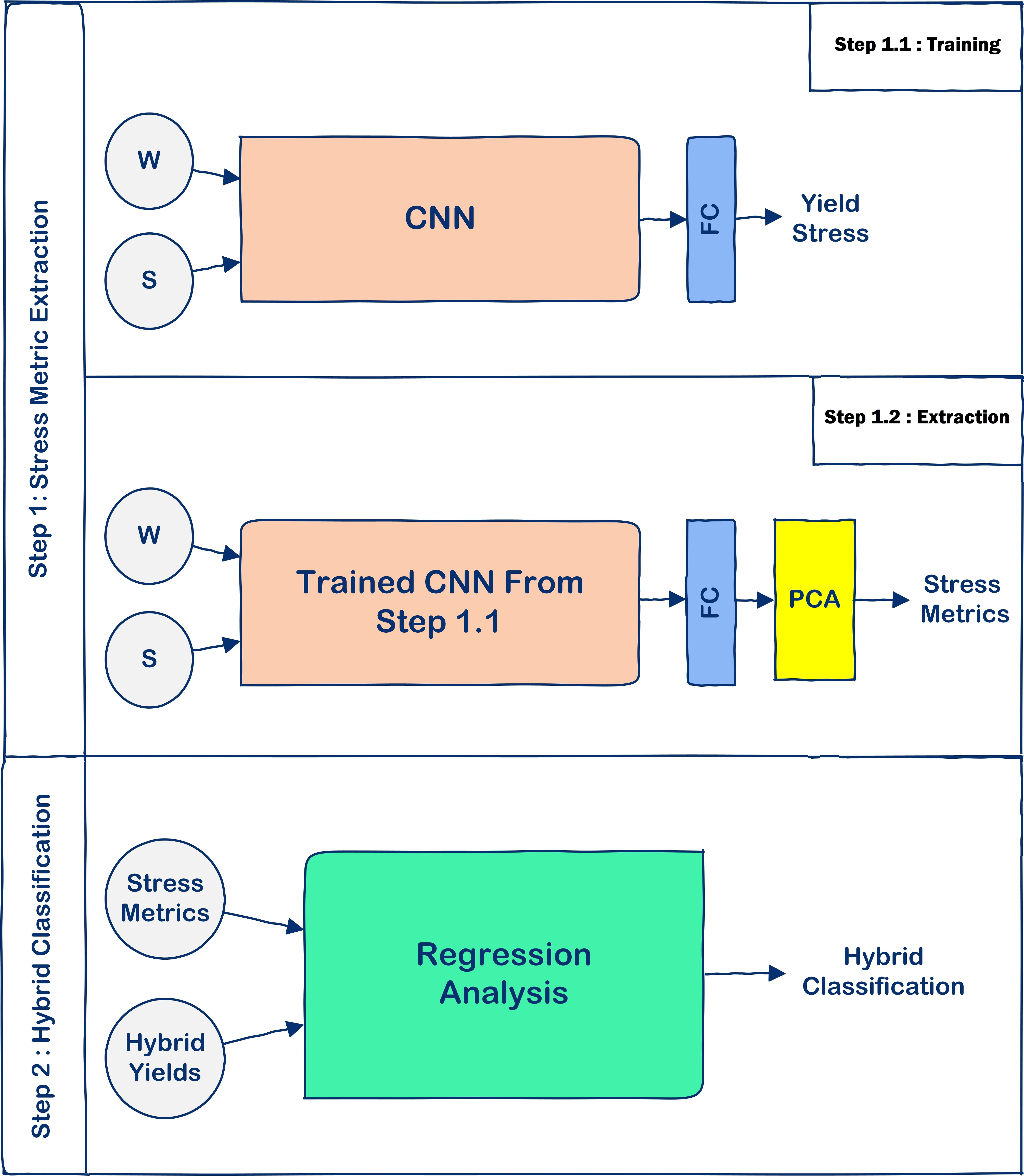}
\caption{Our two-step model for the 2019 Syngenta Crop Challenge, the~goal of which was to develop stress metrics and use these stress metrics to classify corn hybrids as either tolerant or susceptible to drought stress, heat stress and~combined drought and heat stress. W, S, FC and~PCA stand for weather data, soil data, fully-connected layer and~principal component analysis, respectively.}\label{fig:model}

\end{figure}
\unskip

\subsection{Stress Metric Extraction~Method}


We extract stress metric using a deep CNN that was trained to predict yield stress from environmental data. We defined the yield stress in each environment as the amount of yield loss with respect to certain threshold. The~motivation for this definition is that increased yield stress should correlate with decreased yield across environments. Assuming that all hybrids (or at least one of the hybrids with similar genotype) have been tested in sufficiently representative environments, we argue that if the average yield for environment A is less than that for environment B, then environment A is more stressful than environment B. The~yield stress is formally defined as follows. Let $\mu_i$, $\bar{\mu}$ and~$\bar{\sigma}$ denote the average yield for environment $i$, average yield for all environments and~standard deviation of yield for all environments, respectively. Then the yield stress for environment $i$ is defined as
\begin{equation}
\text{max}(0,\bar{\mu}+k \bar{\sigma}-\mu_i),
\end{equation}\label{eq:stress}
where $k$ is a parameter that sets the threshold of yield as the number of standard deviations above the mean. In~this paper, we used $k=3$, which~means that the reference yield is population mean plus three times of standard deviation and~any negative deviation from this yield measures the amount of stress for the environment. This definition is illustrated in Figure~\ref{fig:Growing}.

\begin{figure}[H]
\centering
\includegraphics[scale=0.25]{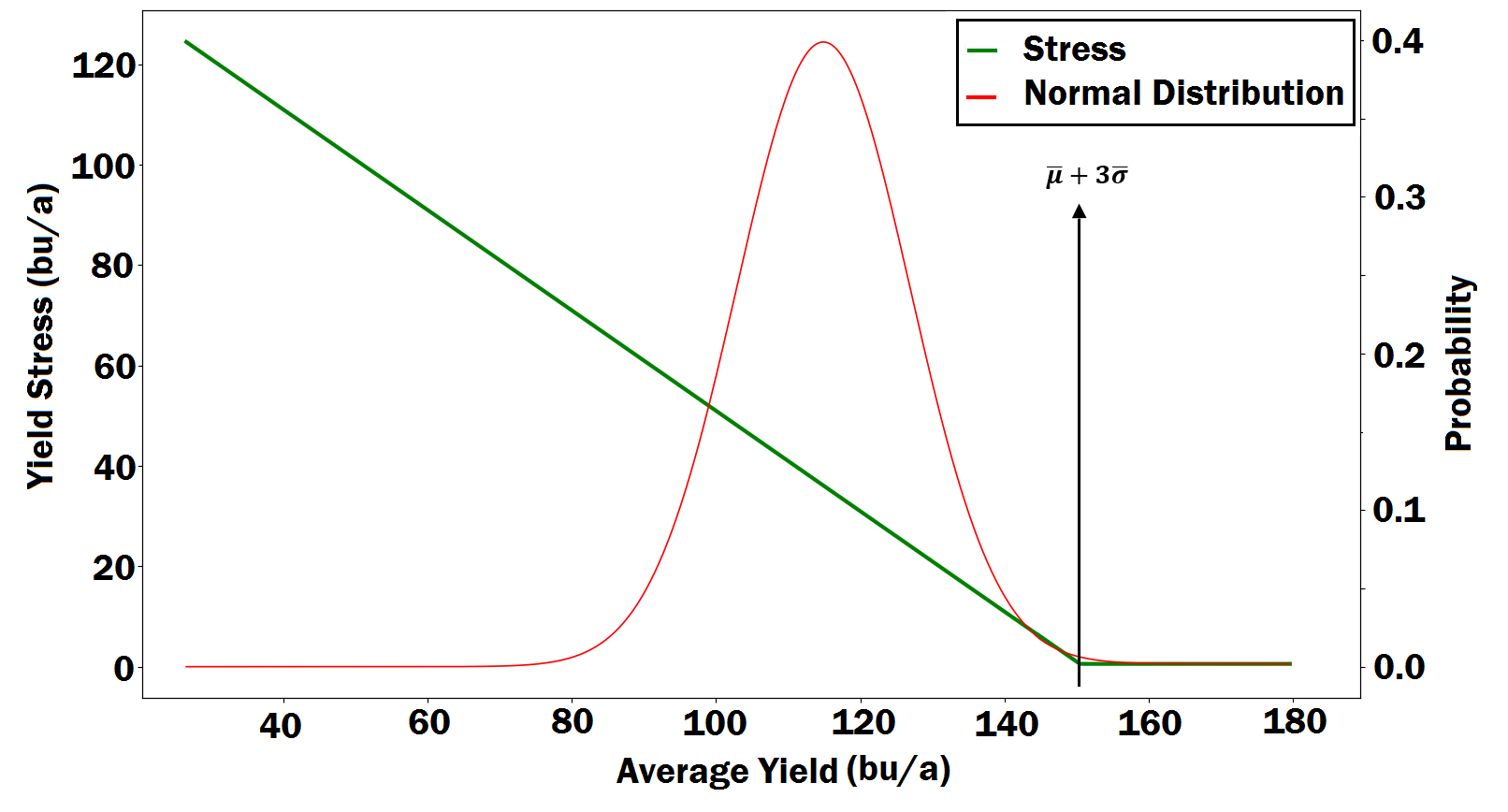}
\caption{Plot of yield stress versus average yield of environments. Yield stress is defined as the amount of yield loss with respect to the threshold of $\bar{\mu}+3\bar{\sigma}$, where $\bar{\mu}$ and $\bar{\sigma}$ are the average and standard deviation of the yield across all environments, respectively. bu/a stands for bushels per~acre.}\label{fig:Growing}

\end{figure}

After defining yield stress, we designed three CNN models to predict the yield stress from soil and weather data, one for each of three stresses—drought, heat and~their combination. Figure~\ref{fig:str} shows the modeling structures of these three CNN models, where ``S'', ``$\text{W}^\text{D}$'' and~``$\text{W}^\text{H}$'' denote environmental variables that belong to the soil, drought and~heat groups, respectively. {The drought group includes day length, precipitation, solar radiation and~snow water equivalent, and~the heat group includes maximum temperature, minimum temperature and~vapor pressure. CNN models also use soil variables, namely field elevation, percentage of clay, silt and sand, available water capacity, soil pH, organic matter, cation-exchange capacity, soil conductivity and~irrigation type as input features. At~the training time, CNN models combine their corresponding input variables to automatically extract necessary features from input data which will be used as stress metrics.}

We designed 1D convolution in the CNN models. The~intuition behind using 1D convolution is to capture the temporal dependencies of weather data and also interaction among weather and soil variables. These temporal dependencies are difficult to capture due to having complex nonlinear relationships~\cite{borovykh2017conditional}. We did not use recurrent-type neural networks~\cite{lipton2015critical} since CNN has a smaller number of trainable weights which can learn temporal dependencies more effectively~\cite{borovykh2017conditional}.

After training three CNNs for predicting yield stress using different combinations of environmental data, we used the output of the FC layer (highest level) of the trained CNN to extract stress metrics, which~is shown in Figure~\ref{fig:emb}.{ The number of stress metrics for each type of stress depends on the number of neurons in the FC layer. For~example, if~the FC layer had $n$ neurons, then there would be $n$ stress metrics for each type of stress, namely heat stress, drought stress, combined heat and drought stress}.  As~the highest level features, these output parameters provide more insightful information than the raw data or lower-level features of the CNN, especially when used as input for other analysis such as classification or regression~\cite{garcia2018behavior,sharif2014cnn, azizpour2016factors,dara2018feature}. { We used these highest level features of the trained CNN models as stress metrics for two main reasons—(1) these features were directly related to the yield stress as a response variable, (2) these features were developed based on the environmental factors (input data) such as weather and soil.} After extracting the stress metrics for all stress types, we~found that some of the neurons of the FC layer were not activated (always zero), so we discarded such stress metrics.{ Since the output of the FC layer is multivariate, we used principal component analysis (PCA)} \cite{wold1987principal}{ to convert the output into a scalar metric for each stress type.} We used PCA for two main reasons—(1) the output of the FC layer was not independent and had high positive correlation with each other, thus using a single metric as a combined metric was reasonable, (2) using one single metric for each type of stress would make the hybrid classification task easier to perform. {Since there were three types of stress, we trained three separate PCA models.  To~train each PCA model, we fed the input data of all environments to the corresponding CNN model and~recorded the output of the FC layer ($n$-dimensional stress metric). Then, we trained the PCA on the extracted stress metrics across all environments.}

\begin{figure}[H]
\centering
\includegraphics[scale=0.35]{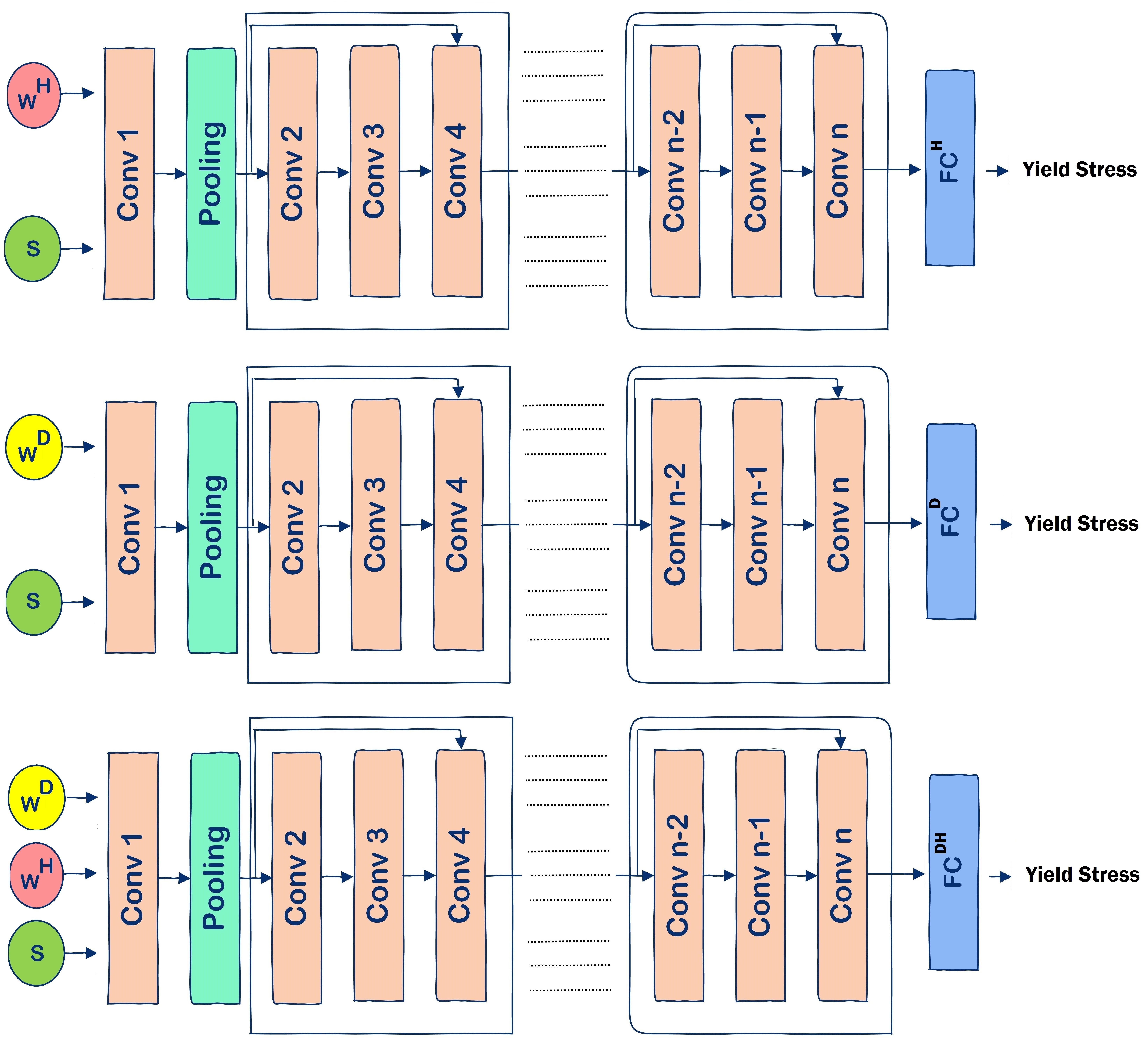}
\caption{The top, middle and~bottom figures show the modeling structures of convolutional neural networks (CNNs) used for heat stress extraction, drought stress extraction and~combined drought and heat stress extraction,~respectively. Every~three layers are grouped into one residual block~\cite{He2016}. $\text{W}^\text{D}$, $\text{W}^\text{H}$ and~S stand for drought related weather data, heat related weather data and~soil data, respectively. }\label{fig:str}

\end{figure}
\unskip


\begin{figure}[H]
\centering
\includegraphics[scale=0.45]{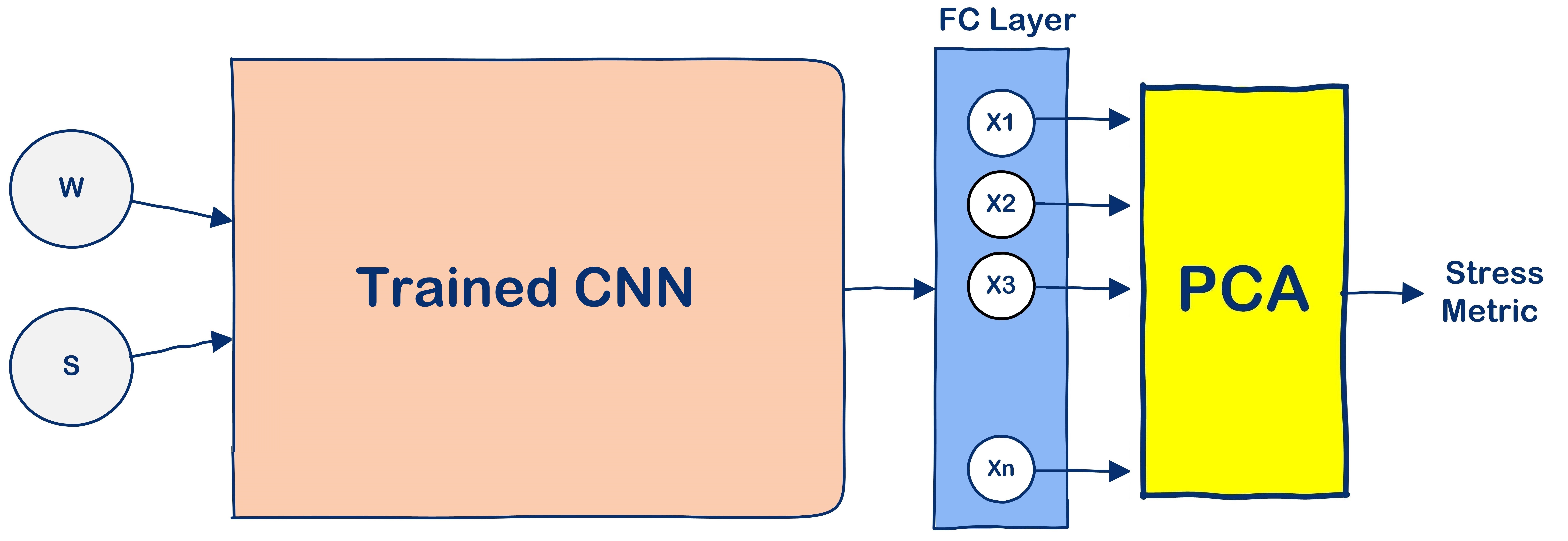}
\caption{Stress metric extraction process after training the CNN model. W and S stand for weather data and soil data, respectively.}\label{fig:emb}

\end{figure}
\unskip

\subsection{Stress Metric Extraction~Results}

We trained our CNN models with the following hyperparameters. All 3 CNN models had the same modeling structure with 13 convolutional layers, followed by the FC layer. Table~\ref{tab:convS} provides the detailed structure of the CNN model. As~shown in Table~\ref{tab:convS}, every 3 convolutional layers construct a residual block in which a shortcut connection was used as in Reference~\cite{He2016}. {The FC layer had 10 neurons}. We tried other deeper or shallower network architectures and found that this network had the best overall performance. We initialized all weights with Xavier method~\cite{Glorot2010}. Batch normalization~\cite{Ioffe2015} was used right after each convolution and before activation for layers 3, 6, 9 and~12. The~rectified linear unit (ReLU) activation was used for all neurons except for the output layer which had linear activation function. We used Adam optimizer with a learning rate of 0.02\% and a mini-batch size of 50. We trained the model for 5000 iterations. We implemented the proposed CNN model in Python using the Tensorflow library~\cite{Abadi2016}.

\begin{table}[H]
  \caption{The table shows the detailed structure of the CNN. FS, NF, S and~P stand for filter size, number of filter, stride and~padding, respectively.\hfill}\label{tab:convS}
\centering
\scalebox{1}{
\def\arraystretch{0.9}

\begin{tabular}{cccccc}
\toprule
 \multicolumn{6}{c}{\textbf{CNN Structure}}\\


 \midrule
 \textbf{Block} &\textbf{Layer Name} &\textbf{FS} &\textbf{NF} & \textbf{S}& \textbf{P} \\
 \midrule
 a & Conv1 &3 &8 &1 &same \\
 \midrule

  - & Average pooling 1 &2 &- &2 &valid \\
 \midrule

 \multirow{3}{*}{b} & Conv2 & 1 & 8 &1 &valid \\ \cmidrule{2-6}

   & Conv3 &3 &8 &1 &same \\ \cmidrule{2-6}

   & Conv4 &1 &8 &1 &valid \\
 \midrule

   - & Average pooling 2 &2 &- &2 &valid \\
  \midrule

 \multirow{3}{*}{c} & Conv5 &1 &8 &2 &valid \\ \cmidrule{2-6}

   & Conv6 &3 &8 &1 &same \\ \cmidrule{2-6}

   & Conv7 &1 &12 &1 &valid \\
 \midrule
   - & Average pooling 3 &2 &- &2 &valid \\
  \midrule

   \multirow{3}{*}{d} & Conv8 &1 &10 &1 &valid \\ \cmidrule{2-6}

   & Conv9 &3 &10 &1 &same \\ \cmidrule{2-6}

   & Conv10 &1 &12 &1 &valid \\
 \midrule

   - & Average pooling 4 &2 &- &2 &valid \\
  \midrule

     \multirow{3}{*}{e} & Conv11 &1 &10 &1 &valid \\ \cmidrule{2-6}

   & Conv12 &3 &10 &1 &same \\ \cmidrule{2-6}

   & Conv13 &1 &12 &1 &valid \\
 \midrule

     - & Average pooling 5 &2 &- &2 &valid \\

 \bottomrule


\end{tabular}
}

\end{table}


The CNN models were trained and evaluated using 10-fold cross validation. Table~\ref{tab:perf} presents the performances of the CNN models on both training and test data with respect to root-mean-square error (RMSE).

\begin{table}[H]

\caption{Yield stress prediction performance of the CNN models using cross validation. D, H and~S stand for drought, heat and~soil, respectively. The~mean$\pm$ standard deviation of the yield stress is 36.31 $\pm$ 22.03. \hfill} \label{tab:perf}
\centering
\scalebox{1}{

\begin{tabular}{lcc}
 \toprule
\hspace{2.2cm} \begin{tabular}[t]{@{}c@{}}\textbf{Model}\end{tabular}  &\begin{tabular}[t]{@{}c@{}}\textbf{Training}\\\textbf{RMSE}\end{tabular}&\begin{tabular}[t]{@{}c@{}}\textbf{Test}\\\textbf{RMSE}\\\end{tabular}\\
 \midrule
CNN using H and S variables &16.65&19.31\\
 \midrule
 CNN using D and S variables &16.46&20.01\\
  \midrule

 CNN using D, H and~S variables &18.38 & 20.69\\
  \midrule

\end{tabular}

}

\end{table}

We demonstrate the effectiveness of our model in stress metric extraction using the following~analysis. We~categorized all environments into high, medium and~low stress based on their observed yield stress defined in Equation~(1), selected 100 environments from each category, fed their corresponding data into the CNN model and~observed whether the output of the FC layer was able to separate environments from three categories. Since the output of the FC layer was high dimensional, we used t-distributed stochastic neighbor embedding (t-SNE) \cite{maaten2008visualizing} method to transfer the high dimensional output of the FC layer to the 2-dimensional space for visualization. The~same process was repeated for all 3 types of stresses using their corresponding CNN models. Results are shown in Figures~\ref{fig:tsne1}--\ref{fig:tsne3}, which~suggest that the proposed CNN models were able to differentiate high, medium and~low stress environments based on their stress~metrics.

\begin{figure}[H]
\centering
\begin{minipage}{.5\textwidth}
  \centering
  \includegraphics[scale=0.19]{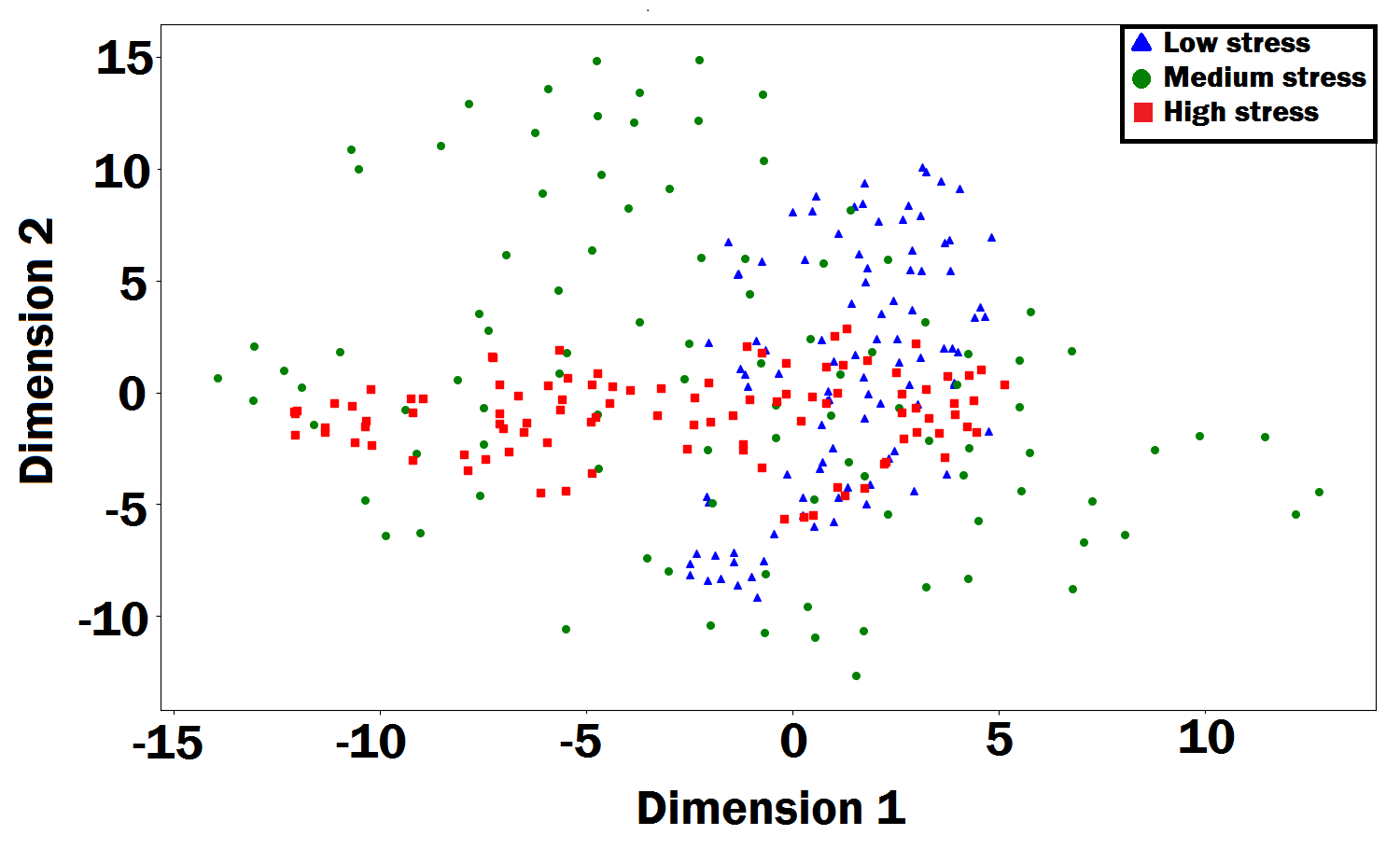}
\end{minipage}%
\begin{minipage}{.5\textwidth}
  \centering
  \includegraphics[scale=0.19]{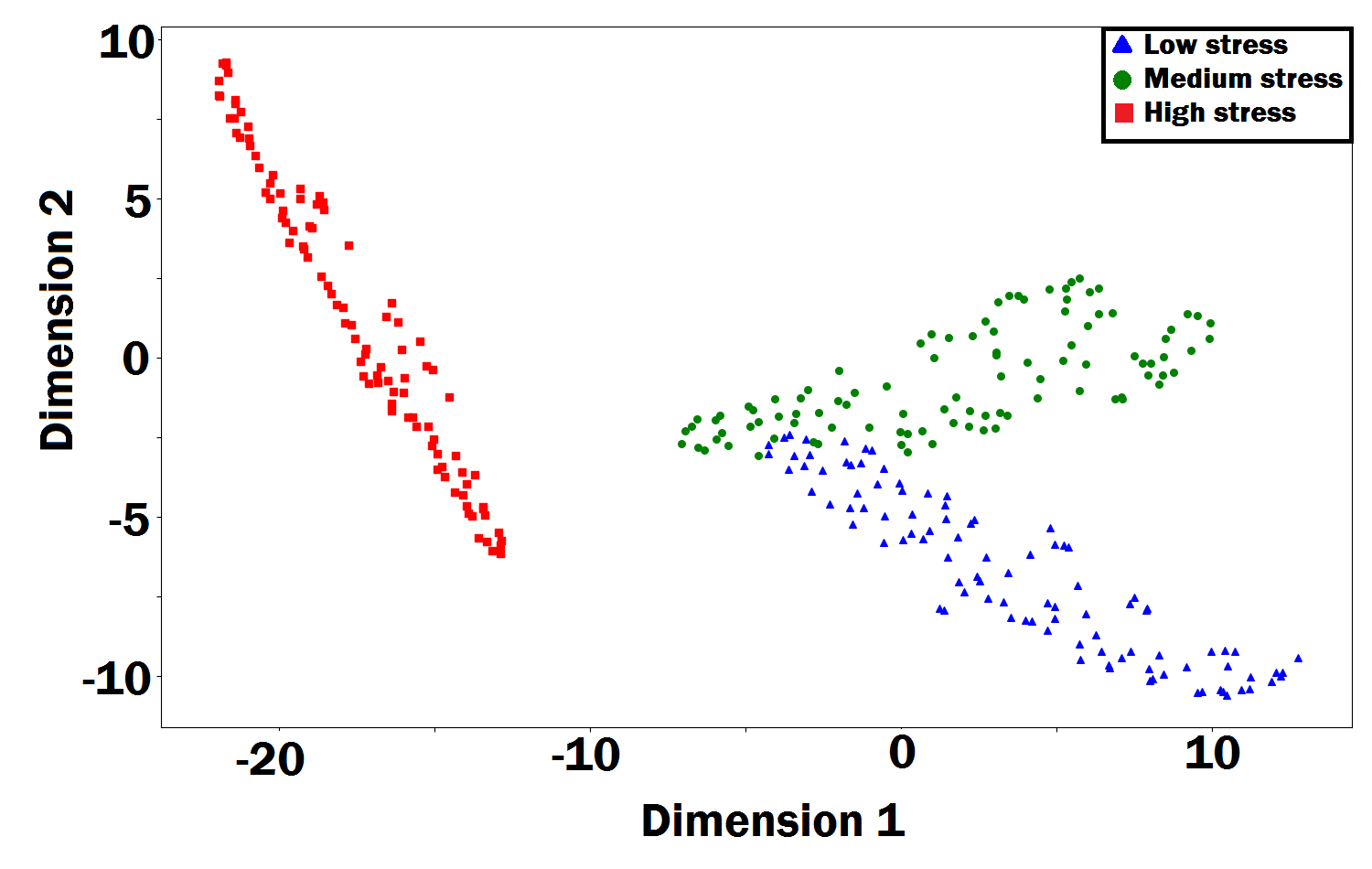}
\end{minipage}
\caption{The plots demonstrate how the CNN model was able to separate environments with high, medium and~low yield stress based on their heat stress metrics.
The left and right plot show the t-SNE embedded output of the heat stress metrics before and after using CNN model, respectively.}\label{fig:tsne1}
\end{figure}


\vspace{-6pt}

\begin{figure}[H]
\centering
\begin{minipage}{.5\textwidth}
  \centering
  \includegraphics[scale=0.18]{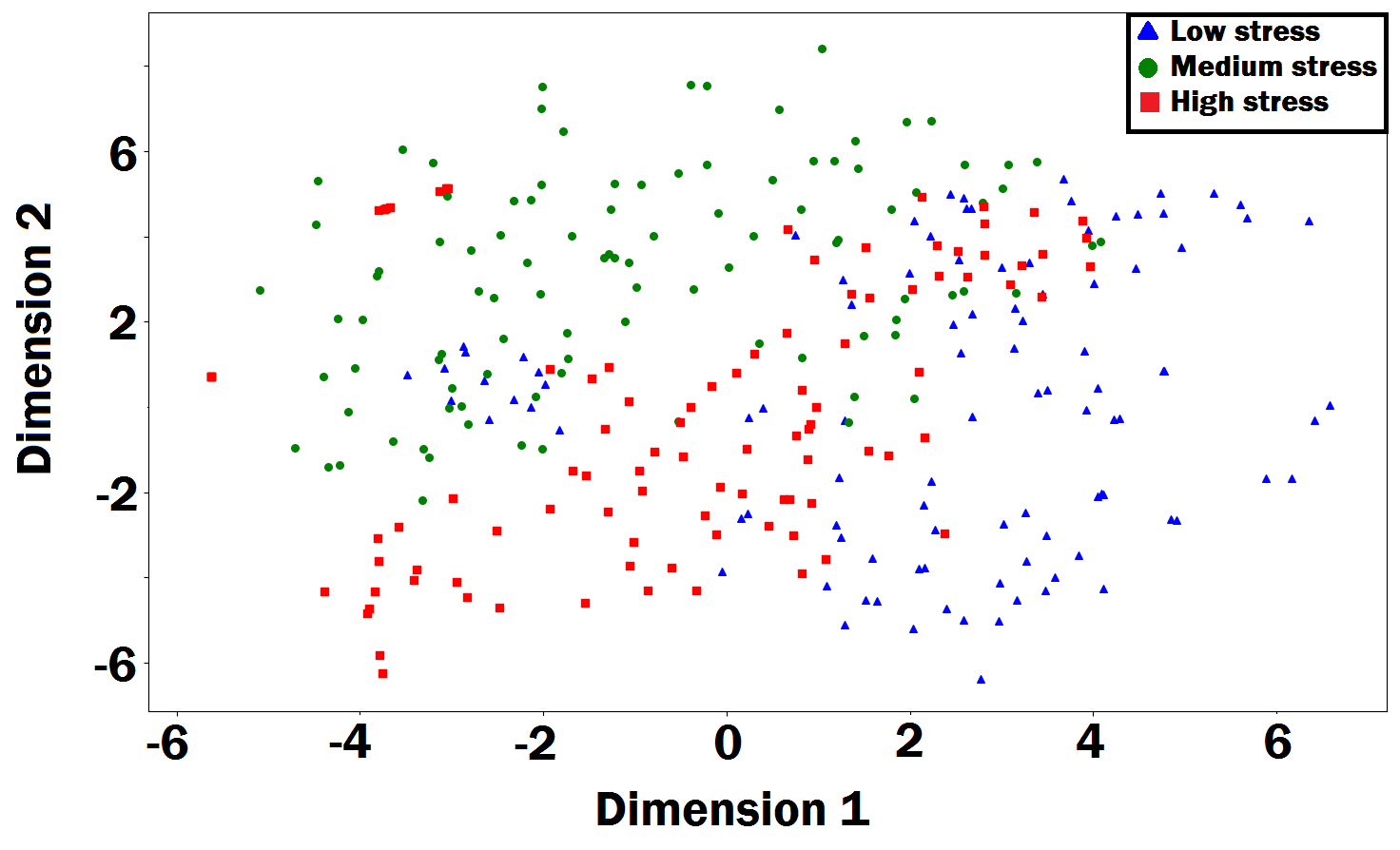}
\end{minipage}%
\begin{minipage}{.5\textwidth}
  \centering
  \includegraphics[scale=0.18]{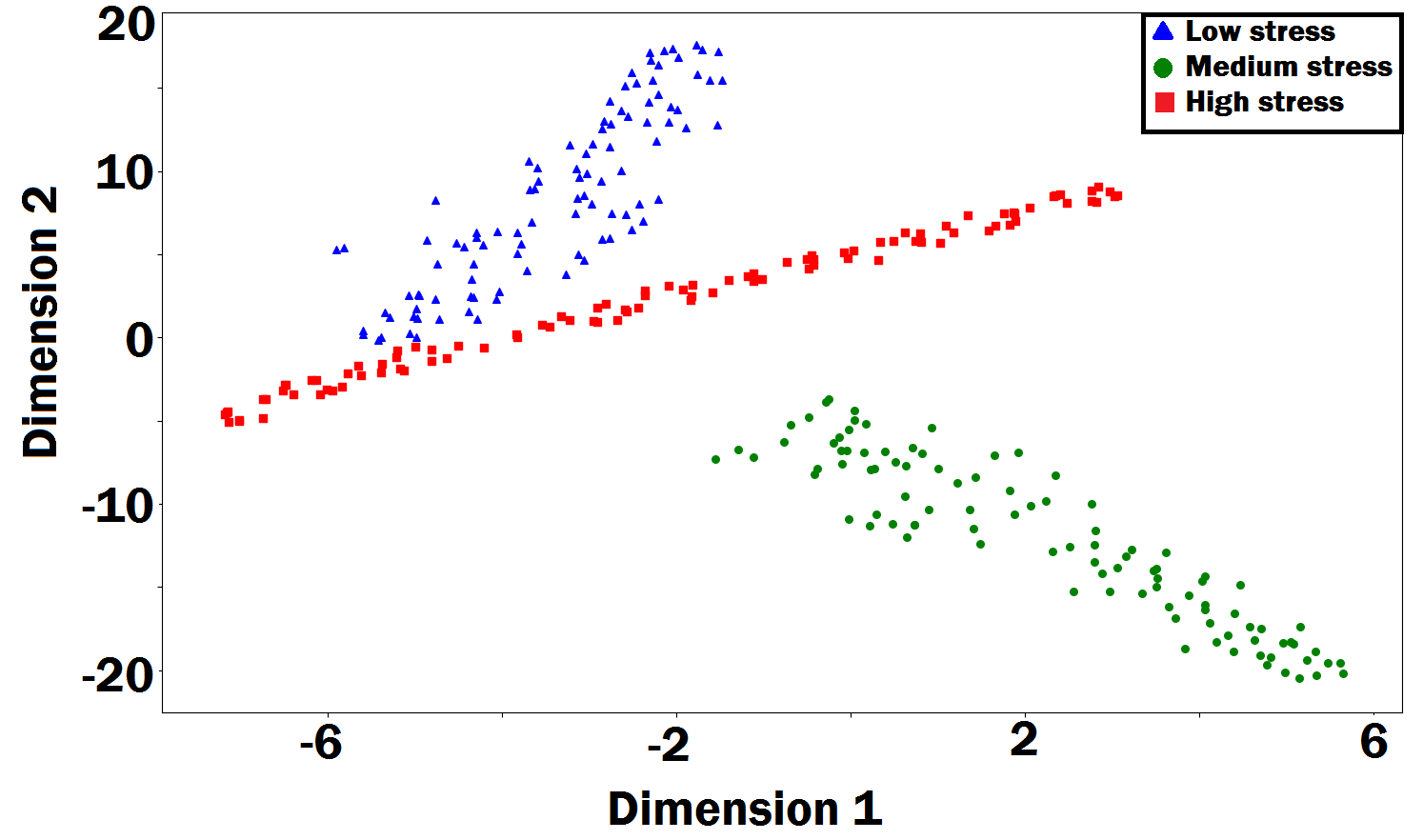}
\end{minipage}
\caption{The plots demonstrate how the CNN model was able to separate environments with high, medium and~low yield stress based on their drought stress metrics.
The left and right plot show the t-SNE embedded output of the drought stress metrics before and after using CNN model, respectively.}\label{fig:tsne2}
\end{figure}


\begin{figure}[H]
\centering
\begin{minipage}{.5\textwidth}
  \centering
  \includegraphics[scale=0.18]{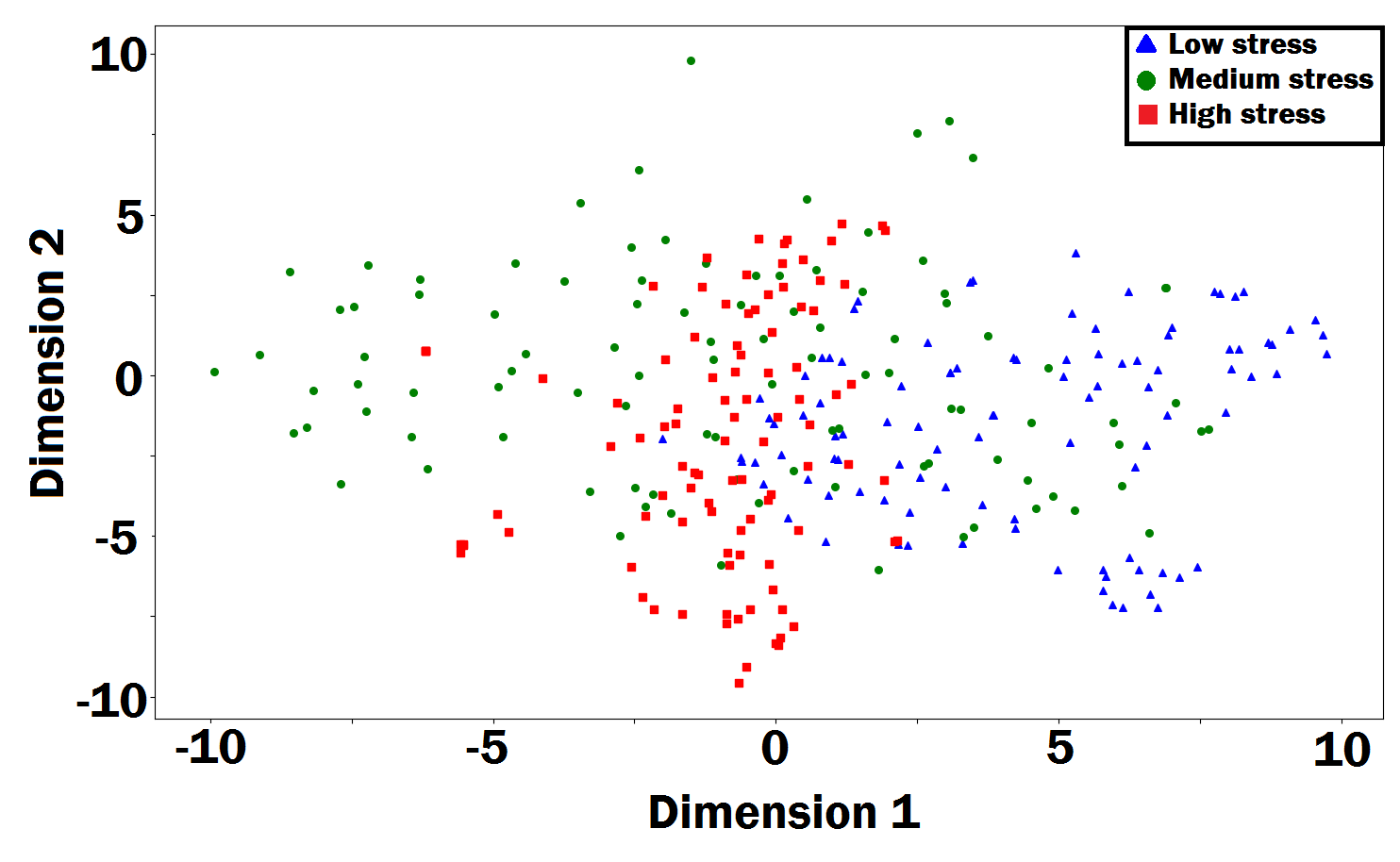}
\end{minipage}%
\begin{minipage}{.5\textwidth}
  \centering
  \includegraphics[scale=0.18]{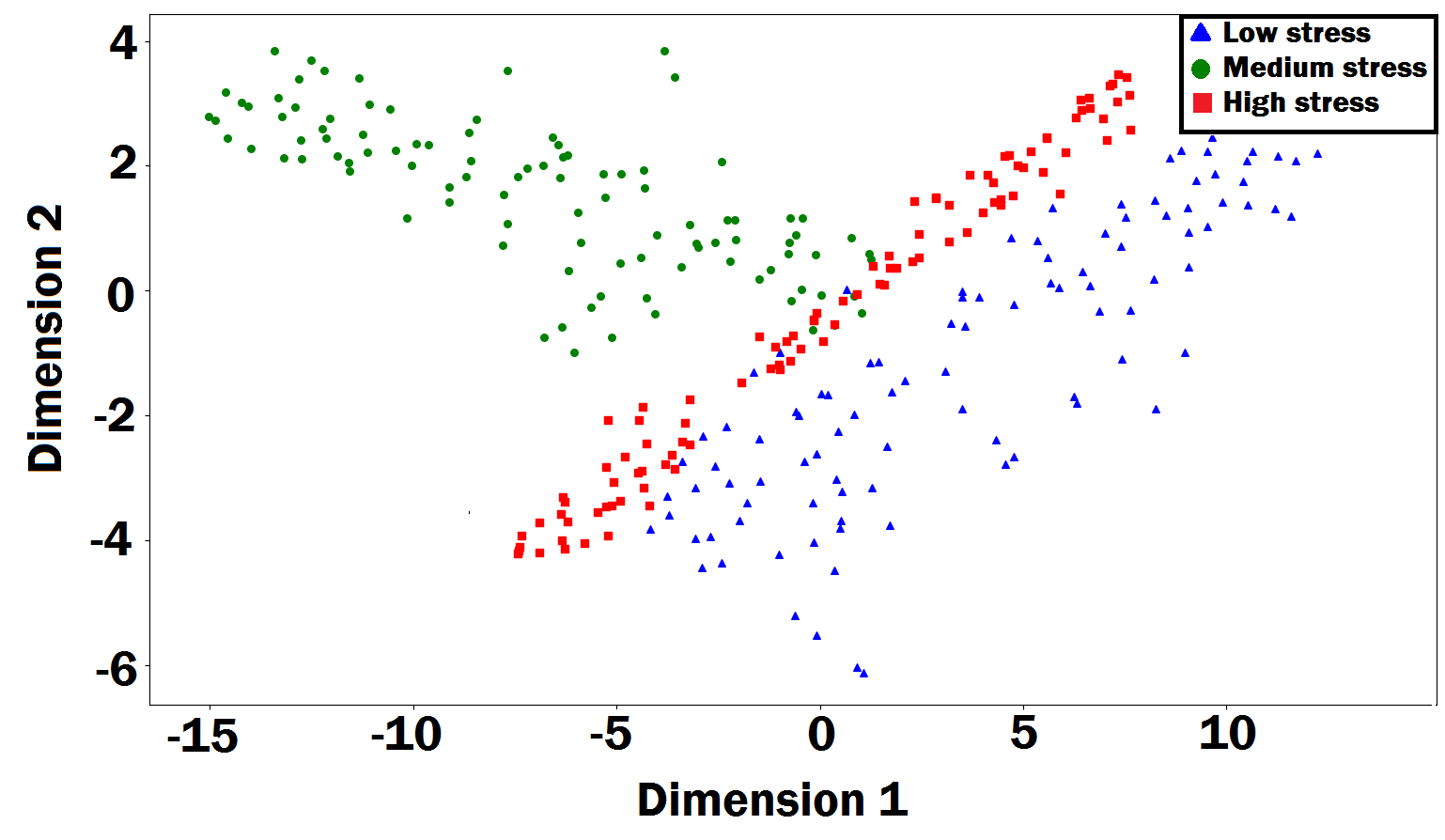}
\end{minipage}
\caption{The plots demonstrate how the CNN model was able to separate environments with high, medium and~low yield stress based on their combined drought and heat stress metrics. The~left and right plot show the t-SNE embedded output of the combined drought and heat stress metrics before and after using CNN model, respectively.}\label{fig:tsne3}
\end{figure}
\unskip

\subsection{Hybrid Stress Classification~Method}

To classify 2452 corn hybrids as either tolerant or susceptible to drought stress, heat stress and~combined drought and heat stress, we performed regression analysis on the yield of hybrids and extracted stress metrics. We conducted linear regression of yield of hybrid against each stress and~classified the hybrid based on the slope of the regression line, since the slope of the regression line indicates the yield adaptability of the hybrid~\cite{10.3389/fpls.2016.01534}.{ Since we applied PCA and converted extracted stress metrics into a 1-dimensional metric for each type of stress, simple linear regression models were used which regressed the hybrid's yields as a response variable on the 1-dimensional stress metric as an explanatory variable}. The~intuition behind this approach is that if a hybrid is tolerant against a type of stress, then the slope of the regression line should be a positive value or a small negative value. Positive or slightly negative slopes indicate that the yield of the hybrid would increase (or~slightly~decrease) under more stressful environments, whereas very negative slopes indicate that the yield of the hybrid would deteriorate significantly under~stress.

In order to classify each hybrid, we first found all environments in which the hybrid was planted and~then we conducted a simple linear regression of the yield of hybrid against each type of stress individually (drought, heat and~combined heat and drought). If~the slope was larger than $-1$, then we classified the hybrid as tolerant to the stress, otherwise the hybrid was classified as susceptible to the stress. Heat or drought stress tolerance was determined independently but~we classified a hybrid to be tolerant to the combined stress only if it is tolerant to both stresses separately, which~means that all three slopes with respect to three stresses must be above $-1$.


\subsection{Hybrid Stress Classification~Results}

We use hybrid H1088 as an example to demonstrate the classification method. The~yield performance under three types of stresses are shown in Figure~\ref{fig:reg} and~the corresponding slopes of the regression lines are given in Table~\ref{tab:reg}. Since this hybrid had consistent yield against drought stress but showed much more sensitivity against heat stress and combined stress, we classified it as drought tolerant but susceptible to heat stress and combined drought and heat~stress.

\begin{figure}[H]
\centering
\includegraphics[scale=0.35]{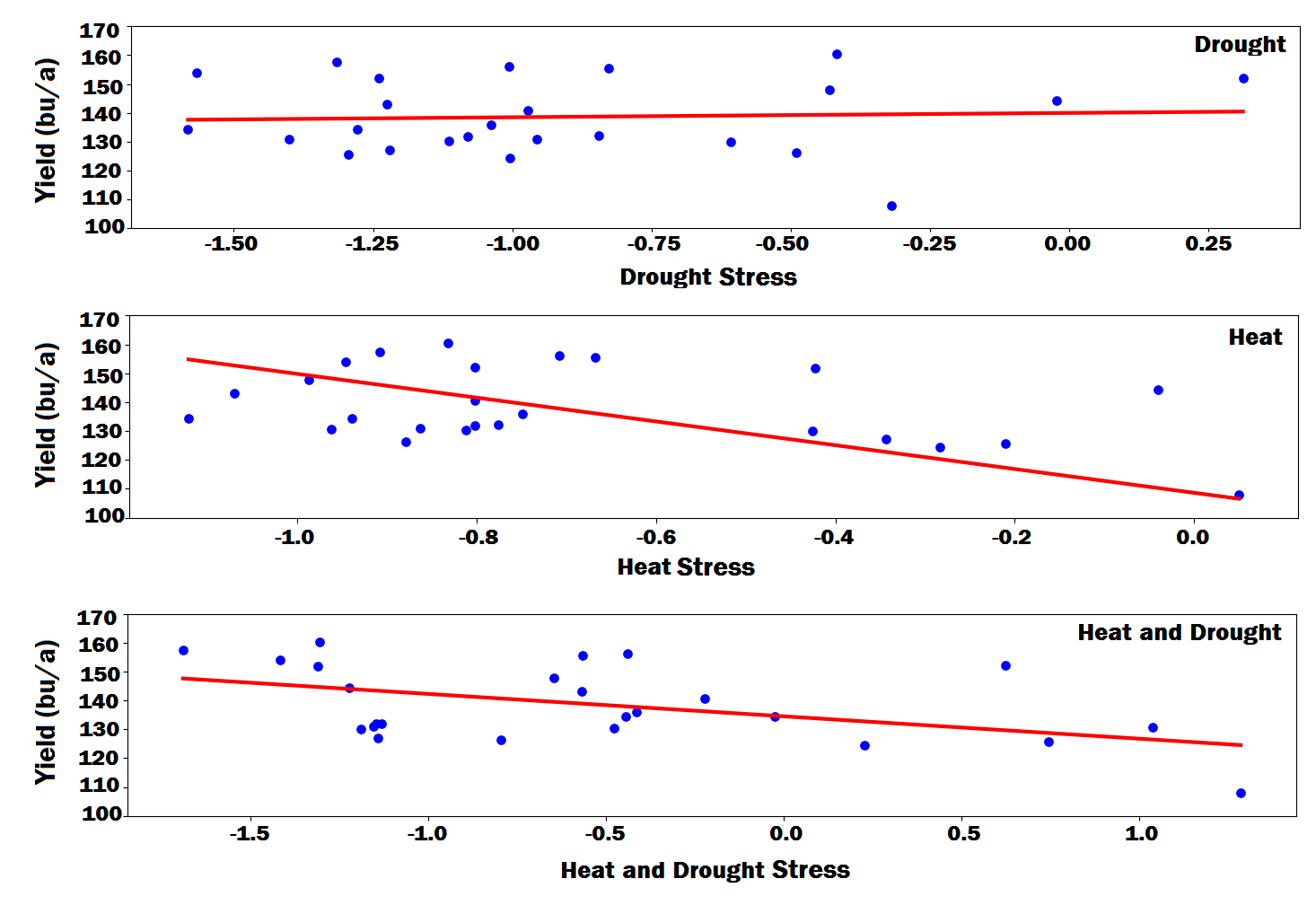}
\caption{The top, middle and~bottom plots show the regression lines for drought stress, heat stress and~combined drought and heat stress for hybrid H1088, respectively. bu/a stands for bushels per~acre.}\label{fig:reg}
\end{figure}
\unskip

\begin{table}[H]
\caption{The regression lines for hybrid~H1088.}\label{tab:reg}
\centering
\scalebox{1}{
\begin{tabular}{ccc}
\toprule
\textbf{Stress Type} & \textbf{Slope} & \textbf{Intercept}\\
\midrule
Drought &1.49&139.99\\
\midrule
Heat &$-$16.21&127.39\\
\midrule
Combined Drought and Heat&$-$7.75&134.46\\
\bottomrule
\end{tabular}
}

\end{table}

Figure~\ref{fig:hist} shows the histograms of slopes for drought stress, heat stress and~combined drought and heat stress for all hybrids. As~shown in Figure~\ref{fig:hist}, the~majority of slopes are less than $-1$, which~reveals that most of the hybrids were susceptible to stresses. Finally, after~doing regression analysis on all hybrids, we found that 121 hybrids were tolerant to drought, 193 hybrids were tolerant to heat and~only 29 hybrids were tolerant to the combined drought and heat stress. Figure~\ref{fig:Venn} summarizes the classification results for all hybrids in a Venn~diagram.

\begin{figure}[H]
\centering
\includegraphics[scale=0.360]{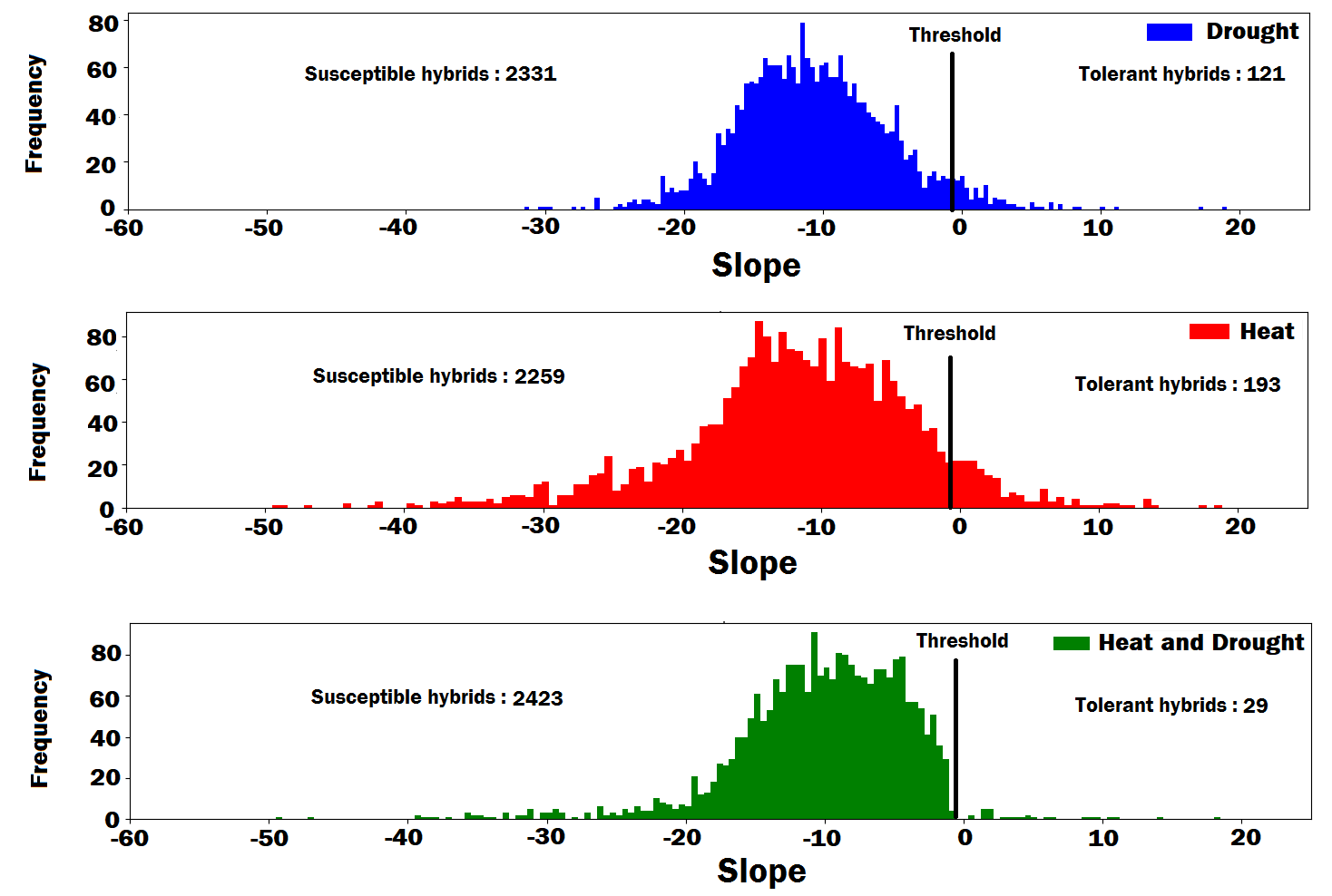}
\caption{The top, middle and~bottom plots show the histograms of slopes for drought stress, heat stress and~combined drought and heat stress for all hybrids, respectively. }\label{fig:hist}
\end{figure}
\vspace{-6pt}

\begin{figure}[H]
\centering
\includegraphics[scale=0.25]{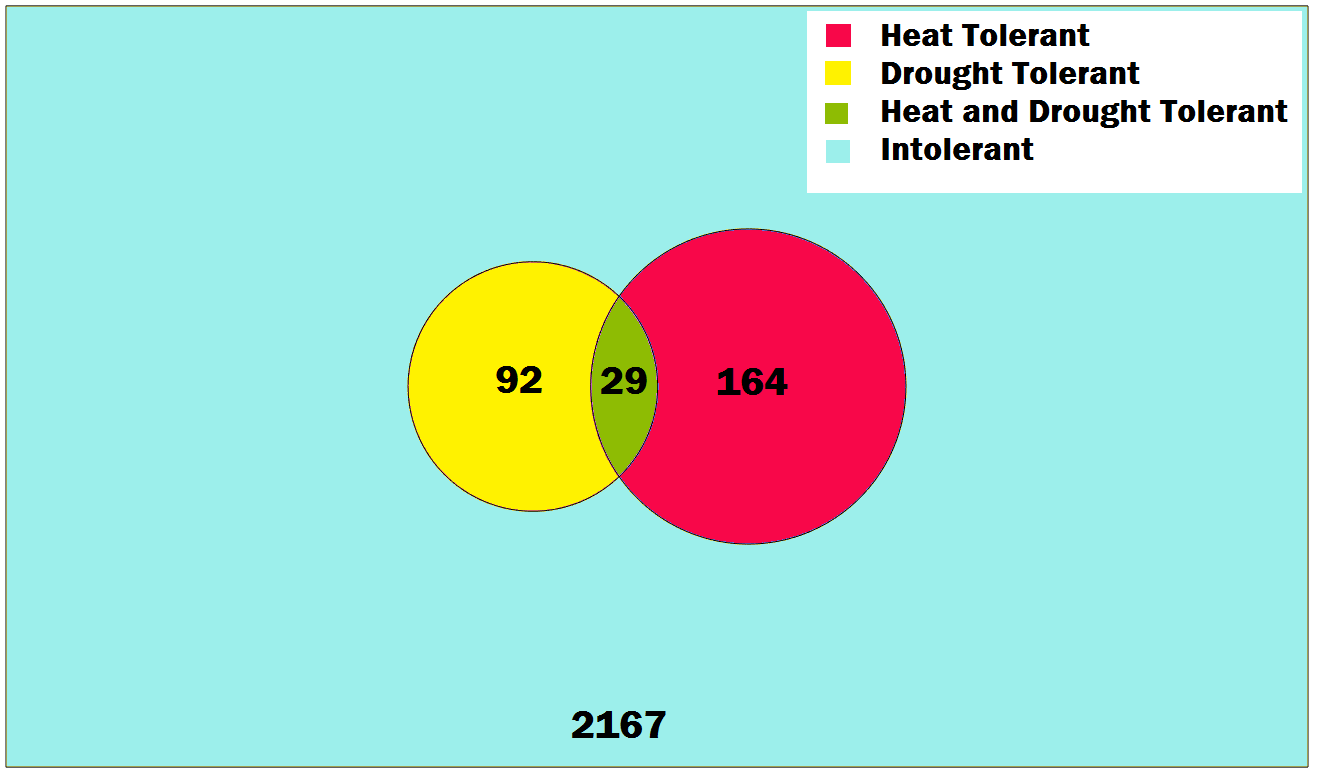}
\caption{The Venn diagram of the hybrid classification~result.}\label{fig:Venn}
\end{figure}
\unskip

\section{Discussion}\label{S4}

In this paper, we presented an unsupervised approach to crop stress classification, which~was based on stress metrics extracted using a deep learning model. This approach consists of two steps. In~the first step, a~deep CNN model was trained to predict yield stress from soil and environment data and historical yield performances of 2452 corn hybrids. Then, we extract stress metrics from the outputs of the trained CNN network, which~implicitly capture the sensitivity of yield stress with respect to heat or drought conditions of the environment. In~the second step, these stress metrics are used to classify the heat and drought tolerance of hybrids using simple linear~regression.

The main contribution of the proposed approach is the novel design of the two-step approach to stress metric extraction and tolerance classification, which~overcomes two major challenges in stress tolerance classification. First, lack of data on the tolerance or susceptibility of hybrids prevents direct application of deep learning models for the classification task. We overcame this challenge by using a deep CNN model to predict the yield stress before extracting stress metrics from the trained model. Yield stress could be calculated from historical yield data, which~allowed the deep CNN model to be trained to approximate the complex relationship between environmental conditions and crop yield stress. The~second challenge that was addressed was the black-box nature of the trained deep CNN model, which~was hard to interpret despite the amount of intelligence hidden within the parameters. We used the PCA technique to extract heat and drought stress metrics to reflect the sensitivity of yield stress with respect to changes in heat and drought conditions. These metrics subsequently allowed all hybrids to be classified as tolerant or intolerant to heat or drought or their combination. As~such, this approach was able to provide an intuitive interpretation of the otherwise unexplainable parameters of the neural networks. Compared with the previous crop tolerance studies, which~mostly used additive linear models to relate the crop yield to weather components such as temperature and precipitation, the~proposed approach used nonlinear models to capture both additive and interaction effects of weather and soil~components.


{In this study, the~stress metrics were developed in the first step of the proposed approach which represented the amount of stress that corn hybrids would face in any particular environment across a growing season. These metrics showed the overall quality of an environment by considering many environmental factors such as weather components and soil conditions. The~results of the stress metric extraction approach were validated using t-SNE method, which~visualized the developed stress~metrics. Computational~results suggested that 2167 hybrids in the data set were intolerant to any stress, 121~hybrids were tolerant to drought stress, 193 hybrids were tolerant to heat stress and~only 29~hybrids were tolerant to the combination of both stresses. Although~only a small subset of hybrids were labeled as stress tolerant according to our results, we also acknowledge that these results are sensitive to the arbitrarily determined thresholds in Figure} \ref{fig:hist},{ which could result in the majority of hybrids being labeled as resistant had their values been set less stringently.} The proposed approach and its classification results were recognized as one of the winners of the Syngenta Crop Challenge by an independent judging committee. {Since we were working with a fixed dataset generously provided by Syngenta and it would be prohibitive to obtain full factorial experimental data for all hybrids planted in all locations, machine learning models are more appropriate than mechanistic ones for this~research.}

\section{Conclusion and Future~Work}

{In this paper, we presented a machine learning approach to identify corn hybrids which were tolerant to heat stress, drought stress and~combined drought and heat stress. The~proposed method solved the problem in an unsupervised way. This approach could be extended to address several future research directions. For~example, all crops would eventually wilt under sufficiently sever stressful conditions, therefore, it might be more informative to measure stress tolerance as a continuous function of the stress than a dichotomous contrast of tolerance or susceptibility. Similar approaches could also be used to study the responsiveness of crops to fertilizer applications, which~would be useful for improving fertilizer efficiency and reducing environmental impact.}

\vspace{6pt}

\authorcontributions{S.K. and Z.K. conceived the study. S.K. implemented the computational experiments. S.K. and L.W. wrote the~paper.}

\funding{This work was partially supported by the National Science Foundation under the LEAP HI and GOALI programs (grant number 1830478).}

\acknowledgments{We would like to express our appreciation to Syngenta and the Analytics Society of INFORMS for organizing the 2019 Syngenta Crop Challenge and providing the valuable datasets. The~source code of the presented method is available on GitHub~\cite{Saeedcode2}. The~data analyzed in this study was provided by Syngenta for 2019 Syngenta Crop Challenge. We accessed the data through annual Syngenta Crop Challenge. During~the challenge, September 2018 to January 2019, the~data was open to the public. Researchers who wish to access the data may do so by contacting Syngenta directly~\cite{s2}. This manuscript has been released as a preprint at~arXiv.}

\conflictsofinterest{The authors declare no conflict of~interest.}



\reftitle{References}

\end{document}